# NEURAL NETWORKS FOR VEHICLE ROUTING PROBLEM

LÁSZLÓ KOVÁCS[1] – ALI JLIDI [2]

***Abstract:*** *The Vehicle Routing Problem is about optimizing the routes of vehicles to meet the needs of customers at specific locations. The route graph consists of depots on several levels and customer positions. Several optimization methods have been developed over the years, most of which are based on some type of classic heuristic: genetic algorithm, simulated annealing, tabu search, ant colony optimization, firefly algorithm. Recent developments in machine learning provide a new toolset, the rich family of neural networks, for tackling complex problems. The main area of application of neural networks is the area of classification and regression. Route optimization can be viewed as a new challenge for neural networks. The article first presents an analysis of the applicability of neural network tools, then a novel graphical neural network model is presented in detail. The efficiency analysis based on test experiments shows the applicability of the proposed NN architecture.*

## 1. INTRODUCTION

### 1.1. Vehicle routing problem and TSP

The Vehicle Routing Problem (VRP) is a very important problem domain in transportation and logistics. It is about organizing vehicle routes efficiently to meet the needs of customers. Basically, VRP involves sending vehicles from certain places to deliver goods to customers and then returning. In literature, many different types of VRP have developed to meet the needs of different industries. These types include options like specific delivery times, sending goods through different stages (from warehouses to local distribution points and then to customers), routes that don't return to the start, handling many kinds of products, using many vehicles, and planning routes for electric vehicles. VRP can be about moving things inside a facility or outside, and it can even cover the whole process of getting products from start to end. As transportation tasks get more complex, with the need to deliver many kinds of products, use different types of vehicles, and deal with various delivery points, using computers to help plan and manage these tasks is becoming very important for being efficient.

The core module of the VRP tasks is to find the shortest path to connect the operation sides. This problem domain is called the Traveling Salesman Problem (TSP) problem. The objective of the TSP family is to discover the shortest Hamiltonian cycle within a weighted graph, where each vertex is visited exactly once. The path's length is determined by the sum of the corresponding edge weights. Solving the problem involves finding an optimal permutation of vertices to create the shortest route. TSP algorithms are versatile and can be applied to various scenarios such as:

- Optimizing transportation routes for delivery and collection tasks [1].
- Heuristically solving knapsack optimization problems [2].

[1] univ. professor, University of Miskolc, Institute of Information Technology
Laszlo.kovacs@uni-miskolc.hu
[2] student, University of Miskolc, Institute of Information Technology
jlidi.ali@student.uni-miskolc.hu

- Planning optimal production sequences [3].
- Addressing printing press scheduling issues [4].

Different variants of the TSP problem exist, including:
- Base Traveling Salesman Problem: a single depot position and a single salesman.
- MTSP - Multiple Traveling Salesman Problem: a single depot position and multiple salesmen.
- MDMTSP - Multi Depot Multiple Traveling Salesman Problem: multiple depots and salesmen, with each salesman assigned to a specific depot.

**1.2. Neural Networks**

The neural network (NN) architecture is the current dominating approach in machine learning to support decision making processes. The NN architecture is a special function approximator which can learn the prediction model from information stored in a training dataset. The main methods in application of NN are the classification and the regression. The input training dataset is given in the form of a list of <X (feature vector), Y (label)> pairs. Each feature vector corresponds to an object instanced in the training space, where the coordinate values represent the different attribute values. In the case of classification, the label denotes the target discrete category, while for the regression problem, the label is the related continuous value.

The main goal of the training process is the build a mathematical model

$$NN: X \rightarrow Y$$

with a good accuracy. The base NN architecture consists of connected neurons organized into layers. Each neuron performs a linear space separation (elementary decision) and generates a non-linear output as the result level of the elementary decision. The generated elementary output values are later compiled into a new hidden feature vector in a high dimensional space. Based on this architecture, the NN can approximate any complex functions. The largest NN systems, like the GPT models for ChatGPT contain more than $10^{12}$ parameters.

The NN architectures are widely used tools for category or value prediction as the model provides the following key properties:
- Non-linearity: Neural networks employ activation functions, enabling them to model complex, non-linear relationships between inputs and outputs.
- Parallel Processing: Neural networks are capable of parallel processing, allowing for efficient computation of large-scale problems across multiple nodes or layers simultaneously.
- Adaptability: Neural networks can adapt and learn from data through training processes, adjusting their parameters to optimize performance on specific tasks.
- Generalization: Trained neural networks can generalize their learned patterns to new, unseen data, making them effective in various prediction and classification tasks.
- Scalability: Neural networks can scale to handle large datasets and complex problems by increasing the number of layers, neurons, or using parallel processing techniques.

- Robustness to Noise: Neural networks can tolerate noisy inputs and incomplete data, often producing reliable outputs even in the presence of uncertainty.
- Representation Learning: Neural networks automatically learn relevant features and representations from raw data, eliminating the need for manual feature engineering in many cases.
- Feature Hierarchies: Deep neural networks can learn hierarchical representations of features, capturing both low-level and high-level abstractions in the data.
- Transfer Learning: Pre-trained neural network models can be adapted to new tasks with minimal retraining, leveraging knowledge learned from previous tasks.
- Interpretability: While neural networks can be complex and opaque, efforts are ongoing to develop techniques for interpreting and understanding their decision-making processes, improving transparency and trustworthiness.

Beside the positive properties, we can also mention the difficulties in the application of NN, like

- Black box model: it is hard or impossible to verify the correctness of the generated model, as the huge complexity is not interpretable by humans.
- It is hard to the find the optimal model as the accuracy on the training set does not ensure the good result on the application dataset.
- The training process can be very time-consuming and expensive.
- The stability of the models is usually not guaranteed.

## 2. ROUTE OPTIMIZATION METHODS

The TSP problem falls into two main domains: the decision problem (DTSP) and the optimization problem (OTSP). DTSP involves determining whether a Hamiltonian cycle with a length not exceeding a given value exists. On the other hand, OTSP aims to find the shortest Hamiltonian cycle. This paper focuses on analysing the OTSP optimization problem, typically formulated as a linear programming problem [5]. OTSP is NP-hard [6], and brute-force algorithms have a time complexity of O(N!). Therefore, heuristic approaches are essential for approximating the global optimum efficiently. TSP heuristic optimization is extensively studied in combinatorial optimization research.

Minimizing the route length holds paramount importance as it significantly im-pacts profitability, directly correlating with factors such as fuel consumption, human resource expenses, the requisite number of vehicles, among others. Thus, route length emerges as a pivotal determinant warranting meticulous consideration. The standard goal of the TSP optimization [7] is to minimize

$$\sum_{i1=1}^{n\_l} \sum_{i2=1}^{n\_l} \sum_{j1=1}^{n\_p} \sum_{j2=1}^{n\_p} \sum_{k=1}^{n\_v} \sum_{t=1}^{n\_t} TDBN_{i1,i2,j1,j2,k,t} \lambda_{j1,j2}^{i1,i2,k,t,m}$$

for all product types *m*, where
    $\lambda_{j1,j2}^{i1,i2,k,t,m}$ is the decision variable and
    $TDBN_{i1,i2,j1,j2,k,t}$ is the distance matrix.

The value of the decision variable is either 1 or 0. In the formulas, k denotes the vehicle index, t is for the time period, *i* is the level index and *j* is the node index.

Various heuristic methods exist for solving TSP problems, including Nearest Neighbour, Nearest Insertion, Tabu Search, Lin-Kernighan, Greedy, Boruvka, Savings, Genetic Algorithm, and Swarm Optimization. A comprehensive analysis and performance comparison of these methods can be found in literature [8][9]10]. Based on this analysis:

- Greedy and Savings methods are the fastest but offer only average tour quality.
- Nearest Neighbour and Nearest Insertion algorithms are outperformed by Greedy and Savings methods in both time and tour quality.
- Optimal route quality can be achieved by employing 3-opt/5-opt methods (Lin-Kernighan and Helsgaun) [11].
- Chained Lin-Kernighan algorithm demonstrates the best overall performance considering both time and tour quality.
- Evolutionary and Swarm optimization methods lag behind k-opt methods in both time and tour quality.
- Tour-merging techniques combined with the Chained Lin-Kernighan algorithm can enhance tour quality but at a higher time cost.

Regarding optimization methodologies, OTSP algorithms can be categorized into two main groups: incremental construction methods and refinement-based methods. Incremental construction methods iteratively add nodes to an evolving optimal route, while refinement-based methods refine an initial route through iterative neighbourhood exploration for local optima.

Heuristic methods typically converge to local optima in the optimization space. The efficiency of these algorithms, in terms of optimal value and time cost, heavily depends on the characteristics of the optimization space, which, in the case of TSP problems, is the space of node permutations. Our study aims to analyse this optimization space to improve approximations of both global and local optima.

## 3. NEURAL NETWORKS FOR TSP

Neural networks have been explored in various ways to tackle the Traveling Sales-man Problem (TSP). The main approaches in application of NN in TSP:

Neural Network-based Heuristics [16]: Neural networks can be trained to act as heuristics, providing approximate solutions to the TSP. These networks take as input the problem instance (e.g., node positions, distances between nodes) and output a permutation of nodes representing a potential solution. Reinforcement learning techniques can be employed to train these networks to generate high-quality solutions through trial and error.

Learning Policies for Optimization Algorithms [13]: Neural networks can learn policies to guide optimization algorithms (such as local search or genetic algorithms) in the search for optimal or near-optimal solutions. The network learns to make decisions at each step of the algorithm based on the current state of the solution, improving the algorithm's effectiveness.

End-to-End Solution Approaches [14]: Neural networks can be used to directly learn mappings from problem instances to solutions without relying on traditional optimization algorithms. These end-to-end approaches often involve training a neural network to approximate the mapping function from problem instances to solutions using large datasets of TSP instances and their corresponding optimal solutions.

Embedding Techniques [15]: Graph embedding techniques, such as graph convolutional networks (GCNs), can be utilized to embed the TSP graph into a continuous vector space where geometric relationships between nodes are preserved. These embeddings can then be fed into traditional optimization algorithms or neural networks to generate solutions.

Attention Mechanisms [15]: Neural networks equipped with attention mechanisms can be used to learn to focus on relevant parts of the input (e.g., node distances) when generating solutions. Attention mechanisms have shown promise in improving the performance of neural network-based approaches for the TSP.

Hybrid Approaches [17]: Combining neural networks with traditional optimization algorithms or metaheuristics can often yield better performance than using either approach alone. For example, neural networks can be used to guide the search space exploration of metaheuristic algorithms or to improve the initialization of optimization algorithms.

Overall, neural networks offer a flexible and powerful framework for solving the TSP, and their effectiveness depends on factors such as problem size, available computational resources, and the specific characteristics of the TSP instances being considered.

### 3.1. Hopfield Neural Network Implementations

The first neural network model for solving TSP was the Hopfield neural network which was presented in 1985 [18]. The network has a recurrent architecture, as it has feedback connections that allow it to store and retrieve patterns from its state. This recurrent nature allows it to exhibit dynamic behaviour over time. Each neuron in the Hopfield network is fully connected to every other neuron, forming a symmetric weight matrix. This connectivity ensures that every neuron influences every other neuron in the network. Considering the operability, the Hopfield network is an energy-based model where each state of the network corresponds to an energy level. The network aims to minimize this energy to reach stable states, which correspond to stored patterns. The weights in a Hopfield network are typically updated using a Hebbian learning rule, which strengthens connections between neurons that fire simultaneously. This learning rule is unsupervised and is based on the principle of association between neurons. In the adaptation of the Hopfiled NN for TSP, the authors defined the following energy function as objective function in the training process Eqs. (5) and (6).

$$\frac{A}{2}\sum_x\sum_i\sum_j X_{xi}X_{xj} + \frac{B}{2}\sum_x\sum_i\sum_y X_{xi}X_{yi} + \frac{C}{2}(\sum_x\sum_i X_{xi}-N)^2 + \frac{D}{2}\sum_x\sum_y\sum_i d_{xy}X_{xi}(Y_{xi'}-Y_{xi'})$$

### 3.2. Graph Neural Network

The other neural network architecture suitable for solving the TSP problem is the Graph Neural Network [19]. Graph Neural Networks (GNNs) are a class of neural networks designed to process data represented in graph structures. GNNs operate on data structured as graphs, which consist of nodes (vertices) connected by edges (links or relationships). This enables them to model relationships and dependencies between elements in the data. GNNs can handle both node and edge features, allowing them to incorporate rich information associated with each node and edge in the graph. These features can be continuous or discrete, and they capture attributes of the elements in the graph. The feature vector describes among others the context of the nodes and edges, i.e. the properties of the neighbourhood elements.

The vector identifies the position of the node using the description of the neighbourhood elements. The feature vectors are sometimes called embedding vectors as they convert the complex neighbourhood into a shorter feature vector.

As the feature vector of a given node depends on the feature vectors of the neighbourhood nodes, we can give that

$$W = f(W).$$

The calculation of the vector components is performed with an iterative process:

$$W_i = f(W_{i-1}).$$

The GNNs architecture typically uses a message-passing mechanism to propagate information between nodes in the graph. At each step, nodes aggregate information from their neighbours, update their own representations, and then pass this updated information to neighbouring nodes. This process is iterated for multiple steps to capture information from distant parts of the graph. The main steps of the calculation of the embedding vectors are the following steps:
- message computation
- message passing to the neighbours
- message aggregations
- message updates

GNNs often utilize attention mechanisms to assign importance weights to neigh-boring nodes or edges during message passing. This allows the network to focus on relevant information and ignore irrelevant or noisy signals in the graph. GNNs can be applied to various tasks such as node classification, graph classification, link prediction, and graph regression. They learn to predict labels or properties associated with nodes, edges, or entire graphs based on their structural and feature information.

The update of the feature vectors can be given with

$$x'_i = W_1 x_i + \sum_{j \in N} \alpha_{i,j} W_2 x_j$$

where the attention parameters are calculated with

$$\alpha_{i,j} = \text{softmax}\left(\frac{(W_3 x_i)^T (W_4 x_j)}{\sqrt{d}}\right)$$

Having the final embedding vectors, the network can be used to perform classification predictions. The input of the prediction NN module is the generated feature vector. The NN module can be used for among others to
- predict a feature of the nodes
- predict a feature of the edges
- predict the existence of nodes
- predict the existence of edges

In order to perform the prediction, the NN must be trained on the corresponding database. In the case of GNN, the training data itself is a graph, usually a subset of the target graph. Thus, for using GNN, we first construct a graph which contains both the training part and the prediction part.

**3.3. GNN for shortest path**

The GNN example presented in [20], presents a GNN network for finding the shortest path between two nodes. The implementation of the GNN is based on the Tensor-flow-GNN library, which includes among others the following operators:
- tfgnn.keras.layers.EdgeSetUpdate
- tfgnn.keras.layers.NodeSetUpdate
- tfgnn.keras.layers.ContextUpdate
- tfgnn.keras.layers.GraphUpdate
- fgnn.keras.layers.MapFeatures

The framework generates training sets from connected graphs which are converted into GNN network. The network stores hidden features for all nodes and edges which denote whether the given component does belong to the shortest path or not. The main goal of the training is to determine the nodes and edges which are the components of the shortest path connecting to given nodes.

The implemented architecture corresponds to the deep GNN model defined in [21]. The generation of the local MLP networks to predict the labels for the edges and nodes, are based on the following algorithm:

```
mlp = tf.keras.Sequential(name="mlp")
for layer_i, size in enumerate(output_sizes):
  layer_activation = activation
  if not activate_final and layer_i == len(output_sizes) - 1:
    layer_activation = None
  mlp.add(tf.keras.layers.Dense(
     size,
     activation=layer_activation,
     use_bias=True,
     kernel_initializer="variance_scaling"
     )
if use_layer_norm:
  mlp.add(tf.keras.layers.LayerNormalization(
     name=f"{name}/layer_norm"))
return mlp
```

As it can be seen the network uses the standard Dense layers. In each iteration of the training, the nodes send the hidden status values to the neighbouring nodes using a message passing mechanism proposed by [22].

For the tested small datasets, the GNN architecture achieved a significant accuracy level (98%). Fig 1. shows some sample graphs from the datasets.

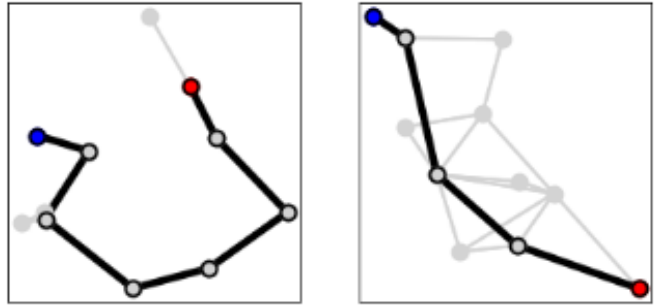

*Figure 1. Sample graphs for the shortest path problem using GNN.*

To adapt the GNN for the TSP problem domain, we updated the network architecture with the following elements:
- we used a fully connected graph, each node pairs are connected
- we used a TSP library [fast_tsp] to determine the optimal cycle in the graph
- there is no start and end nodes in the graph.

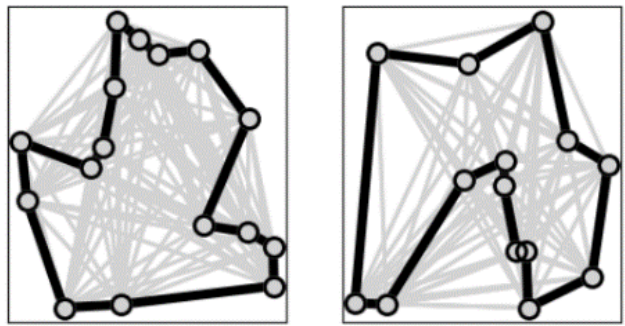

*Figure 2. Sample graphs for the TSP path problem using GNN.*

During the testing of the proposed architecture, we experienced that the resulted model was very weak (20%) on the test datasets. These experiences show that the standard GNN or NN networks are not suitable for direct application on TSP tasks, as
- TSP requires global scale optimization, while message passing is rather a local scale operation.
- The classification in MLP modules performs prediction on single node or single edge level, not at global level.
- Prediction steps in GNN are mainly independent from each other, they do not consider the results in the previous steps.

Thus, because of these experiments, a novel approach should be used for solving TSP with neural network architecture.

### 3.4. Proposed Neural Network Architecture for TSP

In our research project we focused on the development of a special neural network for the TSP problem. In the proposed architecture, the network is aimed as optimizing the length of the cycle route. In order to optimize a global function, we re-structured the architecture of the neural network. In the proposed architecture, the error of the training process is measured not with the usual

$$E = \sum_i (y_i - y'_i)^2$$

difference sum function, but with the

$$E = \sum_i |p_i|$$

where $p_i$ denotes the i-th section of the Hamiltonian cycle. This step requires the development of custom neural network layers which can be implemented with the graph flow model of the Tensorflow framework.

As the usual optimization engine of the neural network architecture, the backpropagation module, implements a gradient descend optimization strategy, the located optimum is usually only a local optimum position. In the different optimization methods, some random relocation steps are involved to give chance of a cost improvement in the undiscovered areas of the optimization space. In the case of genetic algorithm, the following genetic operators are utilized for stochastic steps:
- crossover: exchange of different sections of the feature vectors
- mutation: modification of some section of the feature vector

The proposed architecture contains a modified optimization engine which expands the gradient descend approach with additional crossover and mutation operations. This modification can improve the general efficiency of the embedded backpropagation engine. We remark that one interesting consequence of the structural extension is that we have a unique cost function run during the training, as is shown in Fig. 3.

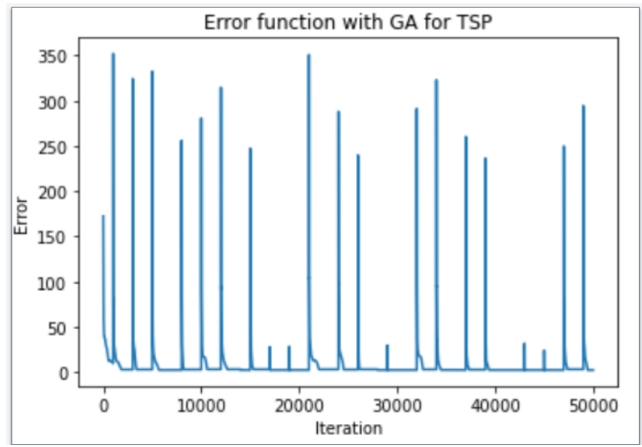

*Figure 3. Error function curve of the updated optimization engine.*

In the curve of the error function, we can see some peaks which mean the points we the search jumped to a new position to start a new gradient descend process. At that step, we move from a local optimum to another position with worse fitness, but later it will find a new local optimum with better fitness.

In order to apply the proposed optimization engine for the TSP problem, we developed a novel representation formalism for the TSP, where integer programming was replaced with a real valued optimization model. In this representation format, the network consists of only simple layers, where the network parameters, edge weights correspond to the relevance level of the graph edges. High relevance means that the corresponding graph edge is part of the Hamiltonian cycle.

## 4. TEST EXPERIMENTS

The implementation of the test environment was developed in the popular Python framework. The neural network objects were built up with Tensorflow-Keras library. In this case, the NN object can be constructed by adding the component layers to the current architecture in a graph flow formalism. The main benefit of this approach is that is more flexible, and it provides more functionality than the usual sequential network model.

```
inputs = Input(shape=(1,))
output = MG_SN_Layer(N_points,Dim,f_modelb,
activation='linear',name="mydense")(inputs)
mg_sn_model = Model(inputs, output)
mg_sn_model.compile(optimizer='sgd', loss='mse')
sprecbn =
CustomCallbackBN(N_points,Dim,Lcalc,Blen
```

In the new model, the calculation of the current cost (error) function depends on the problem domain, thus a custom layer is needed to implement these calculations. As the next code snippet shows, we can use the Tensorflow library to efficiently implement any calculations within the custom layer.

```
def call (self,inputs):
    t1 = tf.constant([0.3,0.7])
    t2 = self.kernel - t1
    t3 = tf.multiply(t2,t2)
    t4 = tf.reduce\_sum(t3,axis=0)
    t5 = tf.constant([1.3])
    t6 = t5 + t4
    return t6
```

Regarding the test results, we can see that this approach provided superior efficiency compared to the previous approaches, it generated very good approximations in the tested cases with small and medium size networks. Figure 4 shows some key steps in the training process to determine the winner cycle.

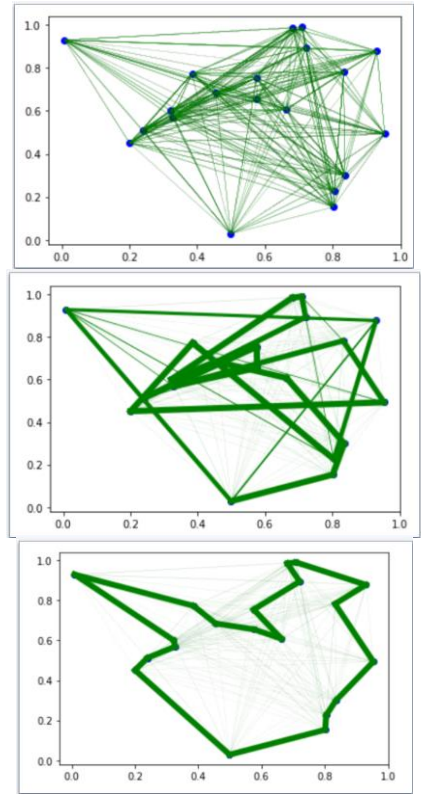

*Figure 4. The shape of the winner cycle in different epochs of the training.*

## 5. CONCLUSION

The VRP and TSP are very important problem domains in transportation and logistics. Due to the complexity of the problem, some kind of heuristics are used to solve the problems of practical sizes. Beside the dominating conventional methods, like the Lin-Kernighan and Helsgaun method, there is an increasing interest in application of so mind of machine learning approaches. In this paper, we investigated the application potential of neural network technology.

In the first approaches of NN-TSP integration, the Hopfield network was the dominating architecture model, it provided the best efficiency. In the current years, the focus of the research turned to the application of the graph neutral network architecture as TSP is defined on graph structure. The performed analysis shows that the standard GNN model has a low efficiency for TSP as it works with local optimization methods while TSP requires a global optimization scope.

As novel approach, we presented an optimization-oriented network architecture which implements a genetic algorithm optimization layer on the top of the standard gradient descend method. The prototype of the proposed architecture was implemented with custom layer elements of Tensorflow framework. The performed tests show that the proposed model dominates the regular GNN-based approaches.

**REFERENCES**

[1] Belhaiza, S., Hansen, P. & Laporte, G. (2014). A hybrid variable neighborhood tabu search heuristic for the vehicle routing problem with multiple time windows. Computers & Operations Research, 52, 269-281, https://doi.org/10.1016/j.cor.2013.08.010

[2] Tavares, J., Pereira, F. B. & Costa, E. (2008). Multidimensional knapsack problem: A fitness landscape analysis. *IEEE Transactions on Systems, Man, and Cybernetics, Part B (Cybernetics)*, **38**(3), 604-616, https://doi.org/10.1109/TSMCB.2008.915539

[3] Chen, H. K., Hsueh, C. F., & Chang, M. S. (2009). Production scheduling and vehicle routing with time windows for perishable food products. *Computers & operations research*, **36**(7), 2311-2319, https://doi.org/10.1016/j.cor.2008.09.010

[4] Christofides, N. (1976). The vehicle routing problem. Revue française d'automatique, informatique, recherche opérationnelle. *Recherche opérationnelle*, **10**(V1), 55-70, https://doi.org/10.1051/ro/197610V100551

[5] Bladimir, T. & Esztergár-Kiss, D. (2023). A review of metaheuristic algorithms for solving TSP-based scheduling optimization problems. *Applied Soft Computing*, **148**, 110908, https://doi.org/10.1016/j.asoc.2023.110908

[6] de la Vega, W. F. & Karpinski, M. (1999). On the approximation hardness of dense TSP and other path problems. *Information Processing Letters*, **70**(2), 53-55, https://doi.org/10.1016/S0020-0190(99)00048-4

[7] Agárdi, A. (2023). *The generalization of the Vehicle Routing Problem: Mathematical model, ontology, algorithms, fitness landscape analysis and applications*. Doctoral dissertation, University of Miskolc

[8] Cordeau, J. F., Laporte, G. & Mercier, A. (2001). A unified tabu search heuristic for vehicle routing problems with time windows. *Journal of the Operational research society*, **52**(8), 928-936, https://doi.org/10.1057/palgrave.jors.2601163

[9] Ai, T. J. & Kachitvichyanukul, V. (2009). A particle swarm optimization for the vehicle routing problem with simultaneous pickup and delivery. *Computers & Operations Research*, **36**(5), 1693-1702, https://doi.org/10.1016/j.cor.2008.04.003

[10] Tunjongsirigul, B. & Pongchairerks, P. (2010). A genetic algorithm for a vehicle routing problem on a real application of bakery delivery. *In 2010 2nd International Conference on Electronic Computer Technology (IEEE)*, 214-217, https://doi.org/10.1109/ICECTECH.2010.5479956

[11] Helsgaun, K. (2015). Solving the equality generalized traveling salesman problem using the Lin–Kernighan–Helsgaun algorithm. *Mathematical Programming Computation*, **7**, 269-287, https://doi.org/10.1007/s12532-015-0080-8

[12] Akyol, D. E. (2004). Application of neural networks to heuristic scheduling algorithms. *Computers & Industrial Engineering*, **46**(4), 679-696, https://doi.org/10.1016/j.cie.2004.05.005

[13] Shi, Y. & Zhang, Y.: The neural network methods for solving Traveling Salesman Problem. *Procedia Computer Science*, **199**, 681-686, https://doi.org/10.1016/j.procs.2022.01.084

[14] Abdel-Moetty, S. M. (2010). Traveling salesman problem using neural network techniques. In *The 7th International Conference on Informatics and Systems* (INFOS) IEEE, 1-6.

[15] Prates, M., Avelar, P. H., Lemos, H., Lamb, L. C. & Vardi, M. Y. (2019). Learning to solve np-complete problems: A graph neural network for decision tsp. *In Proceedings of the AAAI Conference on Artificial Intelligence*, **33**(1), 4731-4738, https://doi.org/10.1609/aaai.v33i01.33014731

[16] Deudon, M., Cournut, P., Lacoste, A., Adulyasak, Y. & Rousseau, L. M. (2018). Learning heuristics for the tsp by policy gradient. *In Integration of Constraint Programming, Artificial Intelligence, and Operations Research Proceedings* 170-181, https://doi.org/10.1007/978-3-319-93031-2_12

[17] Potvin, J. Y. (1993). State-of-the-art survey—the traveling salesman problem: A neural network perspective. *ORSA Journal on Computing*, **5**(4), 328-348, https://doi.org/10.1287/ijoc.5.4.328

[18] Talaván, P. M. & Yáñez, J. (2002). Parameter setting of the Hopfield network applied to TSP. *Neural Networks*, **15**(3), 363-373, https://doi.org/10.1016/S0893-6080(02)00021-7
[19] Zhou, J., Cui, G., Hu, S., Zhang, Z., Yang, C., Liu, Z., ... & Sun, M. (2020). Graph neural networks: A review of methods and applications. *AI open*, **1**, 57-81, https://doi.org/10.1016/j.aiopen.2021.01.001
[20] *GNN for Shortest Path with Tensorflow-GNN*, Retrieved from https://github.com/tensorflow/gnn/blob/main/examples/notebooks/graph_network_shortest_path.ipynb (2024)
[21] Sanchez-Gonzalez, A., Godwin, J., Pfaff, T., Ying, R., Leskovec, J. & Battaglia, P. (2020). Learning to simulate complex physics with graph networks. *In International conference on machine learning,* PMLR, 8459-8468.
[22] Battaglia, P. W., Hamrick, J. B., Bapst, V., Sanchez-Gonzalez, A., Zambaldi, V., Malinowski, M. & Pascanu, R. (2018). *Relational inductive biases, deep learning, and graph networks*. arXiv preprint arXiv:1806.01261.